\documentclass[a4paper]{article}

\usepackage{INTERSPEECH2019}
\usepackage[colorlinks,linkcolor=blue]{hyperref}
\usepackage{stfloats}

\title{Generalized Operating Procedure for Deep Learning: an Unconstrained Optimal Design Perspective}
\name{Shen Chen$^1$, Mingwei Zhang$^2$, Jiamin Cui$^3$, Wei Yao$^3$}
\address{
  $^1$Delta Electronics, Hangzhou, China\\
  $^2$Guahao (Hangzhou) Technology Co.,Ltd., Hangzhou, China\\
  $^3$Key Laboratory of Technology in Rural Water Management of Zhejiang Province, College of Electric Engineering, Zhejiang University of Water Resources and Electric Power, Hangzhou, China}
\email{chenshen@next-aiot.cn, zmw@izmw.me, \{cuijm, yaowei\}@zjweu.edu.cn}

\begin{document}

\maketitle

\begin{abstract}
Deep learning (DL) has brought about remarkable breakthrough in processing images, video and speech 
due to its efficacy in extracting highly abstract representation and learning very complex functions. 
However, there is seldom operating procedure reported on how to make it for real use cases.
In this paper, we intend to address this problem by presenting a generalized operating procedure 
for DL from the perspective of unconstrained optimal design,
which is motivated by a simple intension to remove the barrier of using DL, 
especially for those scientists or engineers who are new but eager to use it.
Our proposed procedure contains seven steps, which are project/problem statement,
data collection, architecture design, initialization of parameters, defining loss function, 
computing optimal parameters, and inference, respectively.
Following this procedure, we build a multi-stream end-to-end speaker verification system,
in which the input speech utterance is processed by multiple parallel streams within 
different frequency range, so that the acoustic modeling can be more robust resulting from 
the diversity of features.
Trained with VoxCeleb dataset, our experimental results verify the effectiveness of our proposed 
operating procedure, and also show that our multi-stream framework outperforms single-stream baseline with 
20 \% relative reduction in minimum decision cost function (minDCF).
\end{abstract}


\section{Introduction}
What is deep learning (DL)? Why it is so popular nowadays? And how to make it for real use cases?
These are some of typical questions with respect to this field, especially for freshmen.
For the first two questions, it is easy to find numerous literature 
introducing this technology \cite{LeCun15-DL,Arulkumaran17-DRL, Guo20-DL3, Zhang20-DLG},
whereas for the last question, there is seldom operating procedure reported on it.
DL is composition of a large number of artificial neurons 
which are assembled into a deep neural network with multiple layers. 
Naturally, it is one kind of representation-learning method \cite{LeCun15-DL}, 
that each layers transform the representations obtained by the former layer(s) into a 
representation at a more abstract and higher level. 
By doing so enough, very complex functions can be learned.
For many years, it has turned out to be very good at solving problems in many domains of 
science, business and government, that have resisted the best attempt of the artificial intelligence community.

Despite the prosperousness of DL at present, how to move forward this technology for 
the future is still deserved to be discussed.
On one hand, continuous efforts on challenging topics, including advanced architecture \cite{Shin18-DNPU, Aggarwal19-MDL}, 
interdisciplinary \cite{Afshar19-FHD, Dai20-DLW}, trustable artificial intelligence \cite{Ajenaghughrure20-RTA, Guo20-EAI}, etc., 
should be made.
On the other hand, more works on how to apply DL to real use cases are also necessary,
so that this technology will not only be manipulated by scientists or engineers in the field of 
computer science, but also can be a powerful tool for those who are interested from other fields, 
and even for all human beings.

Since DL is more and more powerful and is going to be able to do everything
according to the pioneer Geoffrey Hinton \cite{Hao20-APG}, it is meaningful to make this technology
not only powerful, but also easy to operate. 
To this end, we focus on how to operate DL by investigating dozens of 
applications including image recognition \cite{He16-DRL, Huang17-DCC}, 
natural language processing (NLP) \cite{Klosowski18-DLN, Ali20-WER},
automatic speech recognition (ASR) \cite{Park19-SA, Li20-MES}, 
speaker recognition \cite{Nagrani17-VAL, Chung20-IDO}, 
and perceive that these operations are somewhat similar to conventional optimal design without constraints. 
Based on it, we propose a generalized operating procedure for DL from the perspective of 
unconstrained optimal design, expecting to remove the barrier of using it.

Our contributions are as belows:
\begin{itemize}
  \item [1)]
  We propose a generalized operating procedure for DL as a variant of unconstrained optimal design.
  \item [2)]
  We demonstrate our procedure through a task of speaker verification based on VoxCeleb dataset.
  \item [3)]
  We propose a novel multi-stream end-to-end framework with frequency selection technology for robust acoustic modeling.
\end{itemize}
This paper is organized as belows: We review on the related work in Section~\ref{sec:relatedwork}. 
Section~\ref{sec:proposedmethod} describes our proposed operating procedure in detail. 
Section~\ref{sec:experiments} presents our experimental example following the presented procedure.
Section~\ref{sec:conclusion} contains some discussions and conclusion of our work.

\section{Related work}
\label{sec:relatedwork}
Nowadays, it is easy to find abundant resources on DL, as well as lots of guidelines on it.
In \cite{Zhang20-DDL}, an open-source project is presented to make DL approachable, 
in which the concepts, the context and the code are all made available.
As it is declared, the best way to understand DL is to learn by doing, 
this book really provides a starting point on the path to use DL.
In \cite{Quinn19-DDP}, an essential roadmap for designing DL is provided 
with bunches of tips, tools, protocols, and real-world examples.
This easy-to-use guide has everything educators need to construct and 
drive meaningful DL experiences that give purpose. 
In \cite{Stevens20-DLP}, an excellent introduction to PyTorch \cite{Paszke19-PyT} is presented,
which is not only the introduction to PyTorch thorough, but its use in DL is also highly documented and explained. 
In order to realize quick startup for software engineers even without a background in DL,
\cite{Osinga18-DLC} provides lots of recipes written in Python with code available on Github.
including classification or generation of text, image and music.

It can be perceived that most of existing guide on DL contain lots of implementation details 
in form of concepts, tool and code, etc.
Nevertheless, when we first start a real task based on DL, it is not correct to dive into
too many details. Instead, we need to deal with the process as a kind of engineering design problem,
and design the whole process carefully before we implementation.
In engineering field, the finding of a minimum or maximum function problem under some constrained or unconstrained conditions
is usually called optimization problem.
Moreover, almost every engineering design problem can be formulated as optimal design problems,
and DL is also not an exception.
In \cite{Arora04-IOD}, a five-step procedure is proposed for most optimal design problems as follows:
\begin{itemize}
  \item
  Step 1: Project/problem statement. As start up, the statement should be issued on the overall objectives
  of the project/problem, and the requirements to be met. 
  \item
  Step 2: Data and information collection. We need to collect data, material properties, performance 
  requirements, resources limits, and other relevant information.
  \item
  Step 3: Identification/definition of design variables. This step is to identify a set of variables describing
  the system which is so called design variables.
  \item
  Step 4: Identification of a criterion to be optimized. This step is to define a criterion to compare
  the performance of different designs.
  \item
  Step 5: Identification of constraints. As final step, we need to identify all constraints (or restrictions)
  and develop expressions for them.
\end{itemize}

Based on our engineering experience, 
we formulate making DL as unconstrained optimal design problem defined as:
Minimize $f(\vec{\theta})$ without any constraint on parameters of neural networks (NN) $\vec{\theta}$.
To the best of our knowledge, this is the first time to think of the operating procedure for DL.

\begin{figure*}[!bh]
  \centering
  \includegraphics[width=\linewidth]{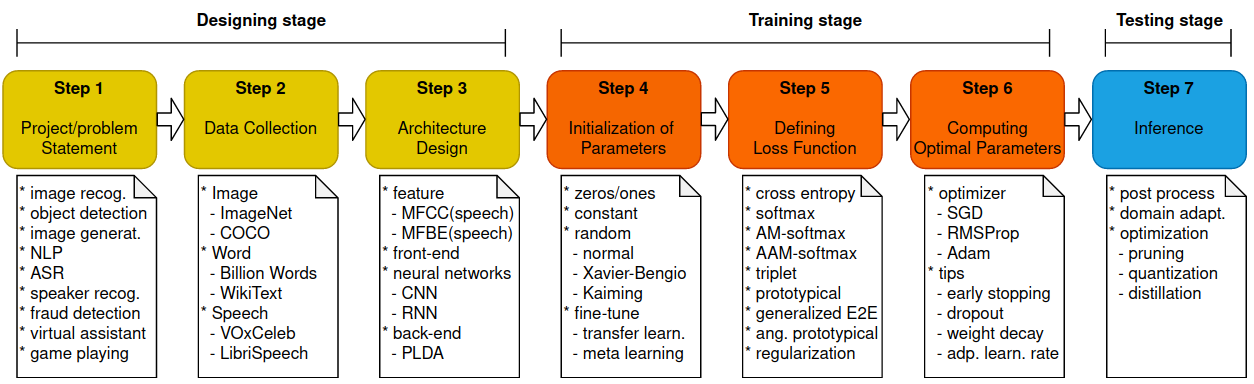}
  \caption{Proposed operating procedure for DL.
  Totally seven steps distributed in three stages.}
  \label{fig:procedure}
\end{figure*}
\section{Proposed operating procedure}
\label{sec:proposedmethod}
Our proposed operating procedure from the perspective of optimal design is illustrated in Figure~\ref{fig:procedure},
which consists of seven steps distributed in three stages.
The designing stage contains step 1 to step 3, which are step 1: project/problem statement,
step 2: data collection, and step 3: architecture design.
The training stage contains step 4 to step 6, which are step 4: initialization of parameters,
step 5: defining loss function, and step 6: computing optimal parameters.
The testing stage is the step 7: inference.
The details of every step are described subsequently.

\subsection{Project/problem statement}
The DL process begins by developing a descriptive statement for the project/problem, 
which is normally done by the owner or sponsor of the project.
More specifically, the statement should be issued on the overall objectives of the project/problem, 
and the requirements to be met. 
\subsection{Data collection}
DL is somewhat a kind of data-driven technology, which usually requires huge amount of 
data for training a model. Fortunately, there are constantly emerging dataset contributed by 
groups or organizations in this area, such as:
\begin{itemize}
  \item
  \textbf{Image.} 
  ImageNet \cite{Deng09-IN} is a large-scale database designed for use in visual object recognition software research,
  which contains 12 subtrees with 5247 synsets and 3.2 million images in total.  
  Others include MNIST, CIFAR, FDDB, COCO, etc.   
  \item 
  \textbf{Word.} 
  Billion Words \cite{Eide16-VAL}, consisting of 1 billion Words is used for benchmarking of language modeling. 
  Others include Text Classification Datasets, WikiText, bAbi, etc.
  \item
  \textbf{Speech.} 
  VoxCeleb \cite{Nagrani17-VAL, Chung18-VDS} is a large-scale dataset for speaker verification obtained from videos in YouTube,
  which contains 153,516 utterances (352 hours in total) from 1,251 celebrities for VoxCeleb1 \cite{Nagrani17-VAL},
  while 1,128,246 utterances (2,442 hours in total) from 6,112 celebrities for VoxCeleb2 \cite{Chung18-VDS}.
  Others include LibriSpeech, CHIME, VoxForge, TIMIT, TED-LIUM, etc.
\end{itemize}
\subsection{Architecture design}
This is obviously the most important step through the whole process, 
and in general, it contains feature, front-end, neural networks (NN) and back-end.
Take speech for example, the commonly used features are Mel Frequency Cepstral Coefficients (MFCC) \cite{Juvela18-SWS} 
and Mel Filter Bank Energies (MFBE) \cite{Li14-FMM}.
The front-end module, or rather feature extractor, is used to extract quality features for subsequent
signal processing by converting the raw input into a relatively lower-dimensional representations.
The NN module is the key component of whole system, 
with CNN and recurrent neural networks (RNN) as two main categories.
CNN has brought about breakthroughs in dealing with non-sequential data such as image and video, 
whereas RNN has shone light on sequential data such as text and speech.
The back-end module is for post-processing after NN.
The most popular approaches for back-end processing contain 
the Probabilistic Linear Discriminant Analysis (PLDA) \cite{Ju19-PLDA}, and neural-based model \cite{Ramoji20-LSS}.

\subsection{Initialization of parameters}
Initialization has a great influence on both the speed of training process and the quality of the ultimate result,
and therefore it requires the right method \cite{Barrera20-RIW}.
Zeros, ones and constant methods initialize NN with zeros, ones and constant value, respectively.
Normal method initializes NN with values drawn from the normal distribution.
Xavier-Bengio \cite{Glorot10-UDT} and Kaiming method \cite{He15-DDR} are two types of advanced random initialization.
Xavier-Bengio method alleviates the problem of gradient vanishing 
by setting the same variance for both the inputs and outputs of each layer.
Kaiming method is similar to Xavier-Bengio method apart from the scaling factor for the weights,
and it takes into account the non-linearity of activation functions, such as ReLU activations.
Moreover, an automated method based on decision trees is presented in \cite{Humbird19-DNN}.
In practice, fine-tune methods such as meta learning and transfer learning, 
are also widely adopted especially when there is lack of target training data.

\subsection{Defining loss function}
Loss function is a method to evaluate how well your model fits your training set.
Cross entropy loss (also called log loss), is used frequently in classification problems.
The softmax loss is modified based on cross-entropy loss, 
which contains a softmax function followed by a cross-entropy loss.
AM-softmax (CosFace) is improved based on softmax loss by normalizing the weights and the input vectors.
By doing so, the posterior probability only relies on the cosine of angle between the weights and input vectors.
AAM-softmax (ArcFace) is similar to AM-softmax, 
except that an additive angular margin penalty is introduced to AAM-softmax.
Triplet loss, prototypical loss, generalized end-to-end (E2E) loss and angular prototypical loss are
some cases of metric learning objectives, 
more details on them are revealed in \cite{Chung20-IDO}.
We consider regularization as part of loss function, because it is essentially functioning
by adding an extra part to original loss so that the distribution of parameters can be regularized.

\subsection{Computing optimal parameters}
This step is to update parameters $\vec{\theta}$ through forward and back propagation.
During training, forward propagation iteratively computes the total loss $\mathcal{L}$ starting from the raw input.
Then, the back propagation will compute the gradient $\frac{\partial{\mathcal{L}}}{\partial{\vec{\theta}}}$,
which is subsequently used to update parameters through gradient descent algorithm.
To do well, optimizer is important, such as stochastic gradient descent (SGD), RMSProp, Adam, etc.
And some tips are also required, including early stopping, dropout, weight decay, adaptive learning rate, etc.
For more details, please refer to Section 11 of \cite{Zhang20-DDL}.

\subsection{Inference}
This step is to make predictions against previously unseen data using the well-trained NN model,
and may need some post process which is not useful for training.
Domain adaptation is usually required due to the domain mis-match between testing and training data.
However, this technology is not restricted to testing stage only, but can also be applied to training stage. 
Moreover, for the reason of meeting requirements of power limitation and trade-off performance in real world,
NN model is often optimized before being deployed for inference \cite{Robins20-TDB}, 
such as pruning, quantization and knowledge distillation.
In \cite{Kozlov20-NNC}, a neural network compression framework supporting various methods, 
is presented for model compression with fine-tuning.

\section{Experiments}
\label{sec:experiments}
\subsection{Operating procedure}
\subsubsection{Project/problem statement}
In our work, we focus on the challenge of recognizing speakers from speech obtained 'in the wild' \cite{Nagrani17-VAL},
aiming to answer the question ``are they from the same speaker?'' with pairwise utterances. 
\subsubsection{Data collection}
We use development part of VoxCeleb2 \cite{Chung18-VDS} dataset for training, which contains 5,994 speakers. 
The VoxCeleb1 test sets (cleaned version) \cite{Nagrani17-VAL} are used as testing sets. 
\subsubsection{Architecture design}
\begin{figure}[!h]
  \centering
  \includegraphics[width=\linewidth]{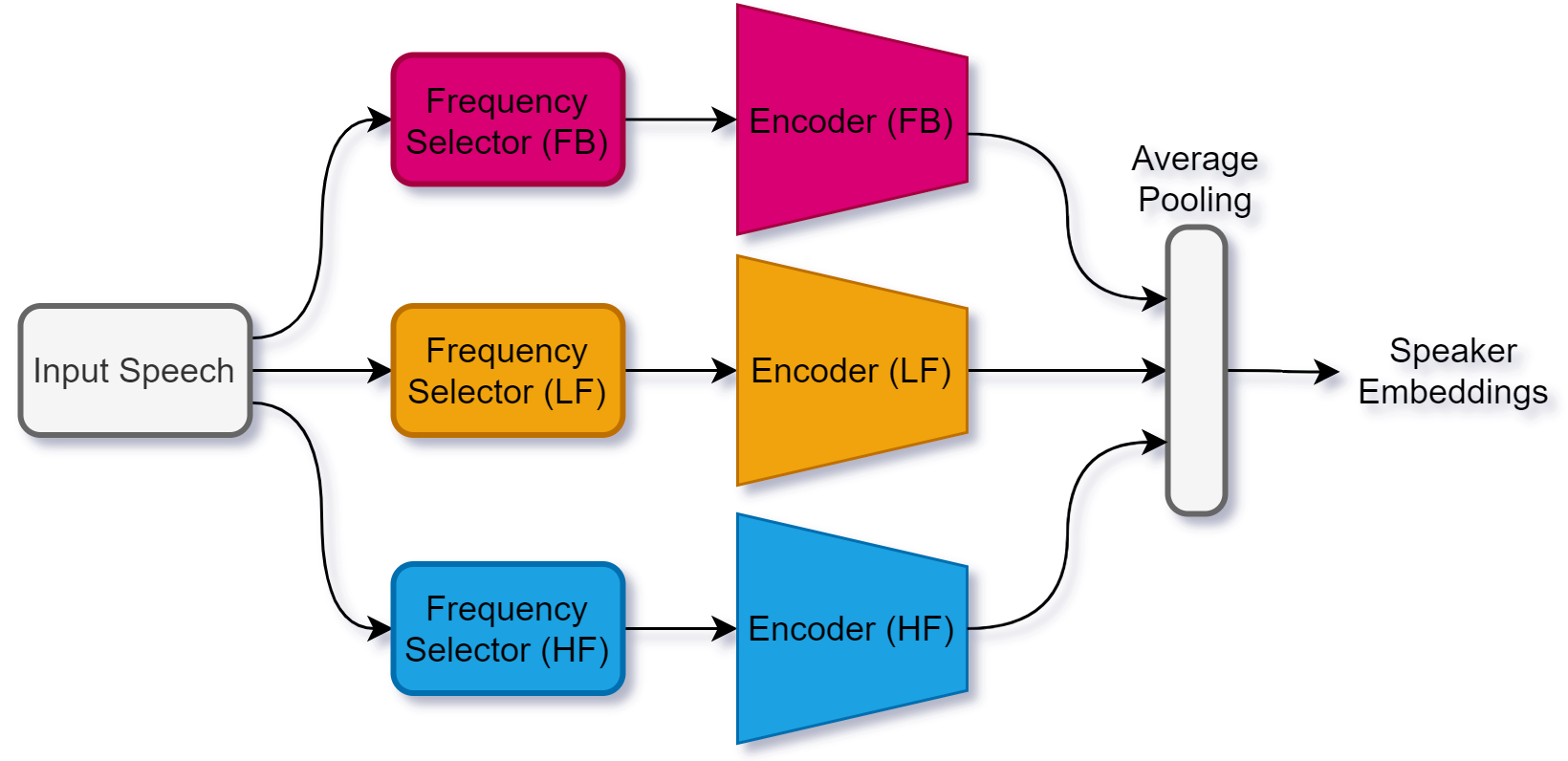}
  \caption{Proposed multi-stream CNN framework for extracting speaker embeddings.
  FB-stream: Full-Band stream, [20, 8000] Hz; LF-stream: Low-Frequency stream, [20, 2000] Hz; 
  HF-stream: High-Frequency stream, [1000, 8000] Hz}
  \label{fig:multistream}
\end{figure}

We use 40-dimensional log MFBE, with a hamming window of 25 ms width and 10 ms step.
The segment length used for training is fixed 2 seconds which is extracted randomly from each utterances.
We build our encoder based on previous work \cite{Yao20-MCN}, 
which is a novel multi-stream CNN framework with frequency selection as illustrated in Figure~\ref{fig:multistream},
and it is available to download at github\footnote{\url{https://github.com/ShaneRun/multistream-CNN}}.

In this framework, each stream has the same network structure but different frequency selector.
The frequency selector is somewhat similar to band-pass filter, 
which determines the frequency range to be listened by encoder. 
The temporal embeddings of each stream is fed into a average pooling layer to generate the final
speaker embeddings.

\subsubsection{Initialization of parameters}
We initialize the parameters of neural networks with Kaiming method as 
described in \cite{He15-DDR}, using a normal distribution.
\subsubsection{Defining loss function}
According to \cite{Heo20-CBS}, the combination of softmax loss and angular prototypical loss
offers improvement over other loss functions. 
For this reason, we adopt it as loss function of our system.
\subsubsection{Computing optimal parameters}
\begin{figure*}[!th]
  \centering
  \includegraphics[width=\linewidth]{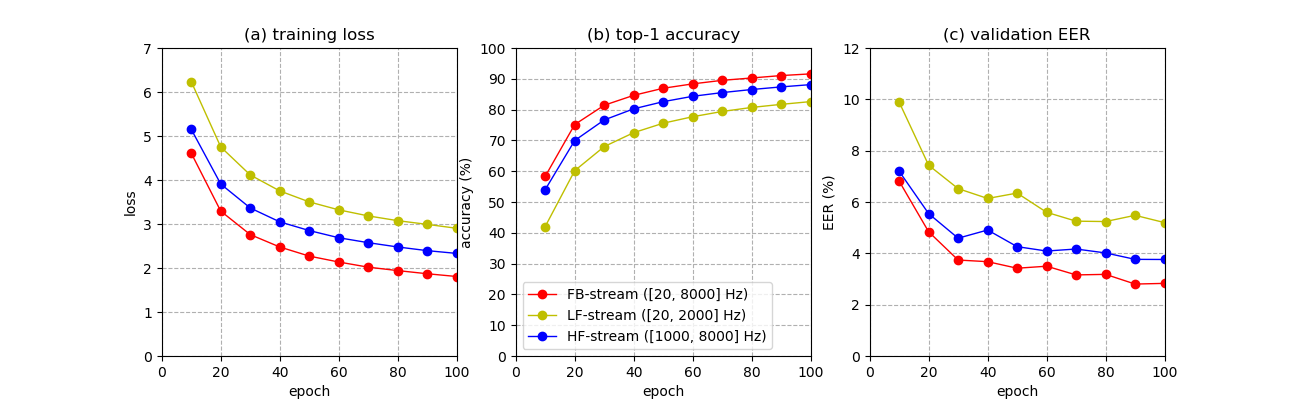}
  \caption{Learning curves.}
  \label{fig:learningcurves}
\end{figure*}

The network is trained in PyTorch using a single GPU (GeForce GTX 1080 TI) with adam optimizer, 
and run for 100 epochs.
Following \cite{Chung20-IDO}, we use a initial learning rate with 0.001, and rate decay of 0.95 every 10 epochs. 
The batch size is set to 400 to make full use of GPU memory (11 GB).   
The learning curves are shown in Figure~\ref{fig:learningcurves}. 
The validation error is printed every 10 epochs during training. 
It can be perceived from the learning curves that system performance becomes better with wider frequency range. 

\subsubsection{Inference}
For inference, the evaluation length is set to 4 seconds, and negative Euclidean distance is used as similarity metric.
No additional post-processing strategy is applied in our system.

\subsection{Results}
The testing results are shown in Table~\ref{tab:example}. 
The multi-stream as highlighted in bold, 
gets 0.168 minDCF and 2.23 \% ERR, which outperforms the single-stream baseline 
with 20 \% relative reduction in minDCF and 18.3 \% relative reduction in EER, respectively.
\begin{table}[!h]
  \caption{Testing results.}
  \label{tab:example}
  \centering
  \begin{tabular}{lllll}
    \toprule
    \textbf{System}  & 
    \multicolumn{2}{c}{\textbf{Vox1-O Set}} &\\
    & minDCF$_{0.05}$ & ERR /\% \\
    \midrule
    FB-stream (baseline)  & 0.210 & 2.73 \\
    LF-stream  & 0.362 & 4.95 \\
    HF-stream  & 0.270	& 3.63 \\
    multi-stream & \textbf{0.168} & \textbf{2.23} \\
    \bottomrule
  \end{tabular}
  
\end{table}

\begin{figure}[!h]
  \centering
  \includegraphics[width=\linewidth]{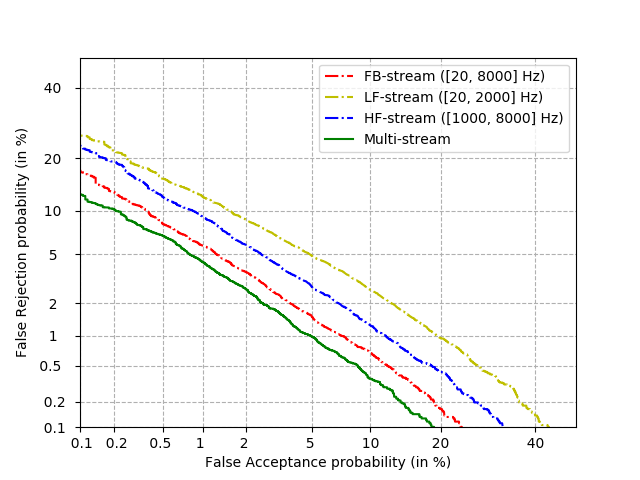}
  \caption{DET Curves.}
  \label{fig:detcurves}
\end{figure}

The Detection Error Tradeoff (DET) curves of FB-stream, LF-stream, HF-stream
and Multi-stream are shown with red dash line, yellow dash line, blue dash line and green solid line respectively
in Figure~\ref{fig:detcurves}.
It can be perceived that our proposed multi-stream framework has comprehensive improvement
compared with our single-stream baseline system.

\section{Conclusions}
\label{sec:conclusion}
In this paper, we propose a generalized operating procedure for DL 
from the perspective of unconstrained optimal design based on our engineering experience.
Furthermore, we verify the procedure through a task of speaker verification.
In our opinion, DL doesn't need to be intimidating,
instead, it can be applied to more interdisciplinary fields, 
and can serve as a powerful tool for those who are willing to use it,
on the condition that they know how to make it with guidance.
However, most of existing guidance provide lots of implementation details,
which is not efficient enough for popularizing DL.
In our work, we intend to provide a graphical architecture of DL process.
Based on this, DL can be made in a more efficient way because of 
optimal design concept and step-by-step operation.
That is the reason why we propose this operating procedure,
and also the meaning of our work by removing the barrier of using DL 
with a generalized operating procedure as guidance.


\bibliographystyle{IEEEtran}

\bibliography{reference} 

\end{document}